# Ontology-Guided Multi-Agent Extraction of Evaluation Objects from Academic Review Texts: Evidence from Chinese Library and Information Science

**Chen, Haolin** Nanjing University, China | 231820024@smail.nju.edu.cn

**Dong, Hongyi** Nanjing University, China | 231820270@smail.nju.edu.cn

**Zhu, Yu** Nanjing University, China | zhu.yu@smail.nju.edu.cn

**Hong, Yijia** Nanjing University, China | 241820194@smail.nju.edu.cn

**Niu, Leiqing** Nanjing University, China | 241820233@smail.nju.edu.cn

**Ye, Jiyuan** Nanjing University, China | yejiyuan@nju.edu.cn

## ABSTRACT

Academic reviews, scholarly commentaries, and book reviews serve as sources of evaluative statements about theories, methods, literature, institutions, and policies, providing valuable evidence for scholarly evaluation. Existing scientific entity extraction methods mainly target research articles and are less effective for evaluation objects, which are often abstract, context-dependent, and characterized by ambiguous type boundaries. This study proposes an ontology-guided multi-agent framework for evaluation object extraction. The framework combines candidate discovery, ontology-constrained classification, and domain review. Experimental results show that it achieves a Precision of 90.33%, Recall of 84.55%, Entity-level F1 of 87.34%, Strict Typed F1 of 79.78%, and Type Accuracy of 91.35%, substantially outperforming rule-based and zero-shot baselines. Ablation results indicate that the multi-agent workflow improves recall and stability, while ontology-based boundary constraints enhance fine-grained classification and reduce category confusion. The framework supports the structured utilization of evaluative scholarly texts and provides methodological support for evidence-based research evaluation and STI mining.

## INTRODUCTION

Scholarly review texts constitute an important genre through which academic communities organize, evaluate, and update knowledge. They encompass review articles, scholarly commentaries, book reviews, and other evaluative texts produced within scholarly publishing and expert assessment contexts (Zou & Hyland, 2020; Blümel & Schniedermann, 2020). Unlike conventional research articles, these texts focus on synthesizing, comparing, and evaluating existing scholarship. A growing consensus in global research evaluation reform holds that scholarly value must be interpreted and validated within specific contexts, expert judgments, and community norms (Leydesdorff et al., 2016). The Leiden Manifesto further emphasizes that quantitative indicators cannot substitute for expert assessment of research quality (Hicks et al., 2015). Against this backdrop, there is an urgent need to identify more appropriate sources of evaluative evidence. Previous studies have shown that review articles contribute to stabilizing knowledge foundations, identifying emerging topics, and conferring academic legitimacy on emerging fields (Blümel & Schniedermann, 2020). Consequently, scholarly review texts are not only vehicles for knowledge dissemination but also rich repositories of evaluative knowledge.

As the scale of scholarly publishing continues to expand, transforming such evaluative texts into searchable and analyzable structured resources has become an important challenge in knowledge organization and scholarly evaluation research. Existing studies have explored the use of the F1000 platform for expert recommendation, open review, and post-publication evaluation, and have examined their relationships with citation-based indicators (Bornmann & Leydesdorff, 2013; Waltman & Costas, 2014). In addition, open peer review practices have provided new objects of analysis for investigating research evaluation processes and reviewer behavior (Thelwall et al., 2021). Nevertheless, a substantial amount of evaluative knowledge remains dispersed across conventional reviews, commentaries, and book reviews, where it is expressed through narrative description, comparison, synthesis, and disciplinary positioning, and has yet to be systematically structured and utilized.

Entity extraction is a fundamental step toward structuring scholarly review texts. Before identifying evaluation dimensions, viewpoints, and relationships, it is first necessary to determine what is being evaluated. However, evaluation objects in review texts are often highly abstract, context-dependent, and characterized by ambiguous type boundaries, making them difficult to identify reliably using surface lexical features or single-pass models. Existing scholarly ontologies, including FRBR, BIBO, VIVO, and CERIF, are primarily designed to support the semantic organization of bibliographic records, scholarly resources, and research activities (D'Arcus & Giasson, 2016; Ding et al., 2010; IFLA Study Group on the Functional Requirements for Bibliographic Records, 1998; Jörg, 2010). However, they lack dedicated representations for identifying evaluation objects and defining their conceptual boundaries in evaluative texts.

To address these challenges, this study proposes an ontology-constrained multi-agent framework for evaluation object extraction from scholarly review texts. The underlying scholarly evaluation ontology consists of three layers: evaluation objects, evaluation dimensions, and evaluation viewpoints. Among them, evaluation objects answer the question of "what is being evaluated" and provide the foundation upon which dimensions and viewpoints are attached. This study focuses on the evaluation object layer, organizing it into four top-level categories—actors, artifacts, abstract entities, and events—and further refining them into a hierarchical type system. Building upon this ontology, candidate discovery, ontology-constrained classification, and domain review are assigned to separate agents, thereby improving extraction recall, type accuracy, and auditability.

## RELATED WORK

### Mining Scholarly Review Texts

Review articles, scholarly reviews, and book reviews are not merely descriptive texts. Rather, they constitute important genres through which academic communities organize research trajectories, formulate judgments, and advance scholarly agendas. Previous studies have shown that review articles help portray the state of a field, identify future research directions, and contribute to knowledge production, while scholarly book reviews are characterized by the dual requirement of "objective description–evaluative negotiation," in which evaluation is not supplementary information but the core of the genre (Blümel & Schniedermann, 2020; Tang, 2012). Therefore, the value of such texts lies not in simply describing content but in the scholarly judgments and knowledge-organizing cues embedded within them.

Existing studies have primarily employed citation analysis, co-word analysis, and topic modeling to investigate scholarly communication and knowledge structures. While effective for revealing macro-level patterns, these approaches provide limited support for identifying evaluation objects, relationships, and dimensions embedded in review texts (Hjørland, 2013; Leydesdorff & Nerghes, 2017).

Although computational methods have been applied to evaluative texts, existing studies mainly focus on general review scenarios such as product and online reviews (Carenini et al., 2006; Kraychev & Koychev, 2012). Academic evaluation texts differ substantially because their objects are often abstract knowledge entities, including theories, methods, disciplines, and scholarly debates. Therefore, specialized object-type frameworks and extraction approaches are required for structuring evaluative knowledge in scholarly reviews.

### Entity Extraction from Scientific Texts

The continuous growth of scientific literature has driven the development of information extraction technologies. Early studies primarily captured semantic associations through word embedding learning (Pennington et al., 2014), while the emergence of deep pre-trained language models advanced contextualized semantic modeling for information extraction (Devlin et al., 2019). For specialized domains, researchers further developed domain-specific language models such as SciBERT and BioBERT, achieving substantial improvements in named entity recognition and relation extraction tasks (Beltagy et al., 2019; Lee et al., 2020). Meanwhile, research scenarios have gradually expanded to encompass long documents, multiple entities, and complex relationships, promoting the evolution of scientific knowledge extraction toward document-level understanding and structured knowledge discovery (Jain et al., 2020).

However, for long-form scientific literature, challenges related to entity boundary identification, type classification, and cross-document semantic understanding have increasingly constrained extraction performance. Numerous studies have demonstrated that domain-specific shifts in vocabulary distributions constitute a major obstacle to model generalization, as representations learned from general-purpose corpora often fail to adequately capture specialized knowledge contexts (Lee et al., 2020). In cross-domain named entity recognition tasks, even with domain-adaptive pre-training, models remain prone to confusion between fine-grained entities and higher-level concepts (Liu et al., 2021). Many critical entities and their relationships are distributed across different sections of a document, requiring the integration of full-text context for accurate identification, while existing models still exhibit a substantial gap from human annotation in document-level reasoning capabilities (Jain et al., 2020). Consequently, the primary bottlenecks in entity extraction are no longer limited to text representation capabilities but increasingly involve the delineation of conceptual boundaries, the utilization of knowledge constraints, and the modeling of complex semantic relationships.

Although scientific literature mining has been widely applied to knowledge discovery and topic analysis, evaluative scholarly texts remain underexplored. Review articles and related scholarly texts play an important role in scholarly evaluation and disciplinary sense-making, yet existing research has focused primarily on their communicative functions and document characteristics rather than their internal evaluative knowledge structures (Spink et al., 1998; Blümel & Schniedermann, 2020). Recent computational studies mainly address topic discovery, citation analysis, and concept normalization, providing limited support for identifying higher-level evaluation objects and judgments embedded in scholarly reviews (Marrone et al., 2022). Consequently, a substantial research gap remains in fine-grained knowledge organization and entity extraction for evaluative academic texts.

### Ontology-Guided and Multi-Agent Extraction

Ontology-Based Information Extraction (OBIE) introduces formalized and explicitly defined conceptual specifications to provide semantic constraints and knowledge guidance for the extraction process, thereby improving the accuracy, consistency, and interpretability of extraction results (Wimalasuriya & Dou, 2010). Ontologies can explicitly define the core concepts, properties, and relationships within a domain, providing a unified semantic framework for knowledge organization and machine reasoning (Gruber, 1993; Noy & McGuinness, 2001). In the field of scholarly information organization, existing ontologies have been successfully applied to bibliographic description, citation relationship modeling, research community representation, and research activity management, but they provide limited support for modeling evaluative knowledge, such as stance expressions, scholarly judgments, and support–challenge relations.

However, existing scholarly ontologies are primarily designed to support the structured organization of papers, authors, institutions, and citation networks, making them insufficient for representing stance expressions, critical judgments, and support–challenge relationships embedded in evaluative texts. Although previous ontology research has enabled the representation of document entities, citation contexts, and citation counts, it has paid relatively little attention to evaluative statements, scholarly judgments, and their semantic dependencies. Consequently, a substantial gap remains in the modeling of evaluative knowledge (Di Iorio et al., 2014; Gangemi et al., 2006; Wang et al., 2020). At the same time, single-model approaches are prone to boundary drift, type confusion, and factual instability in complex extraction tasks. Research on multi-agent systems has demonstrated that role specialization, iterative discussion, and structured verification can significantly improve the stability and consistency of complex tasks (Hong et al., 2023; Du et al., 2024).

In conclusion, despite progress in scholarly text mining, information extraction, and ontology-guided extraction, limited attention has been paid to the knowledge structures embedded in evaluative academic texts. In particular, systematic typologies of evaluation objects and ontology-guided multi-agent extraction frameworks remain largely unexplored in Chinese academic review texts. This study addresses this gap.

## METHODOLOGY

### Academic Evaluation Ontology

#### *Theoretical Foundations*

The purpose of constructing the Academic Evaluation Ontology in this study is to provide stable semantic boundaries for extracting evaluative knowledge from academic review articles, scholarly commentaries, and book reviews. Unlike general bibliographic ontologies or research information management ontologies, the primary concern of this study is not how publications, authors, and institutions are connected, but rather what entities are evaluated in evaluative academic texts, from which dimensions they are evaluated, and what judgments are ultimately formed. This approach is consistent with the classical definition of an ontology as an "explicit specification of a shared conceptualization" and aligns with the fundamental role of ontologies in providing semantic constraints and type boundaries for information extraction tasks (Gruber, 1993; Noy & McGuinness, 2001; Wimalasuriya & Dou, 2010).

The ontology design draws upon formal ontology, knowledge organization theory, and research on scholarly entity modeling. Guarino's (1998) distinction between Endurants and Perdurants provides the ontological foundation for the top-level category structure. The entity hierarchy principles of FRBR/LRM inform the modeling of Artifact-type entities, including publications, data resources, and normative documents (IFLA Study Group on the Functional Requirements for Bibliographic Records, 1998). Ranganathan's (1967) faceted classification theory guides the design of Organization entities, thereby avoiding the conflation of institutional forms and functional roles. The theories of Kuhn (1962) and Hjørland (2002, 2013) provide the conceptual basis for abstract knowledge objects such as theory, model, framework, discipline, and school of thought. In addition, the entity model of OpenAlex serves as a reference for ensuring compatibility between the ontology and existing scholarly knowledge infrastructures (Priem et al., 2022).

#### *Ontology Structure*

The present study adopts an evaluation-oriented ontology modeling approach, treating evaluative activities as the core semantic units of scholarly review texts and organizing the Academic Evaluation Ontology into a three-layer structure: Evaluation Object, Evaluation Dimension, and Evaluation Viewpoint. This section distinguishes the complete ontology from the experimental task addressed in this study. The full ontology comprises three layers: Evaluation Object, which answers the question of what is being evaluated; Evaluation Dimension, which specifies the aspects from which the evaluation is conducted; and Evaluation Viewpoint, which captures the judgments being made. Since evaluation objects serve as the foundational anchors to which dimensions and viewpoints are attached, this study focuses on the Evaluation Object layer and validates its effectiveness.

Evaluation Objects define the entities being evaluated, such as theories, models, methods, disciplines, institutions, policies, and scholarly turns. Evaluation Dimensions specify the aspects along which evaluations are conducted, including theoretical contribution, methodological innovation, scholarly impact, normativity, applicability, and limitations. Evaluation Viewpoints record evaluative statements together with their polarity, intensity, and supporting evidence, such as affirmation, questioning, revision, comparison, supplementation, and criticism. Among these modules, Evaluation Objects function as the semantic anchors for dimensions and viewpoints. Only after the evaluation objects are reliably identified can subsequent dimension recognition and viewpoint extraction be meaningfully grounded in explicit referents.

Furthermore, the Evaluation Object layer is subdivided into Agent, Artifact, Abstract Entity, and Event. This classification also draws upon formal ontological distinctions among different modes of existence: agents, artifacts, and abstract knowledge objects remain relatively stable over time, whereas events such as scholarly turns, academic controversies, and policy initiatives exhibit clear processual and temporal characteristics (Guarino, 1998). This distinction helps avoid treating ontologically heterogeneous entities—such as scholars, publications, theories, and scholarly movements—as simple parallel categories.

### *Evaluation Object Layer*

The Evaluation Object layer is organized around four top-level classes: Agent, Artifact, Abstract Entity, and Event. This structure provides a stable, interpretable, and verifiable classification framework for the extraction task. At the implementation level, the ontology is further refined into 4 L1 classes, 14 L2 subclasses, and 59 L3 leaf types. This hierarchical structure enables the system to perform coarse-grained categorization before progressively moving toward fine-grained classification, thereby reducing the complexity of individual classification decisions and facilitating subsequent error analysis and ablation studies. Compared with a flat label schema, a hierarchical ontology is better suited to handling the numerous neighboring concepts and ambiguous boundaries commonly found in evaluative academic texts.

The ontology organizes evaluation targets into four top-level classes:

· Agent, actors involved in the production, organization, dissemination, or governance of scholarly activities, including scholars, practitioners, and organizations.

· Artifact, human-produced outputs and resources that can be read, used, retrieved, or executed, such as publications, datasets, standards, policies, and software.

· Abstract Entity, intellectual constructs independent of a specific material carrier, including concepts, theories, models, frameworks, methodologies, methods, disciplines, and schools of thought. This represents the most challenging category in the present extraction task.

· Event, temporally bounded scholarly or institutional processes, such as debates, research programs, policy initiatives, developmental stages, and academic meetings.

### *Boundary Rules and Annotation Principles*

Evaluative academic texts contain numerous object types that are semantically similar yet ontologically distinct, such as methodology, method, and technique; theory, model, and framework; and organizations with different functional orientations. To improve annotation consistency and classification interpretability, this study formalizes key epistemological boundaries as explicit classification rules and applies them to both manual annotation and automatic extraction. By incorporating these boundary constraints into the ontology, the system can maintain consistent semantic standards during entity recognition and type assignment, thereby reducing type drift and boundary ambiguity.

To reduce ambiguity among semantically adjacent categories, five predefined confusion groups were incorporated into the ontology:

1. Methodology / Method / Technique: Methodology refers to an overarching research stance or approach, Method to a reusable research procedure, and Technique to a specific operational procedure. Specific operations should not be classified as Methodologies or elevated to complete Methods.

2. Research / Service / Professional / Governance Organization: Organizations are classified according to their primary function in knowledge production, service provision, professional community support, or governance activities, rather than by name or institutional affiliation.

3. Theory / Model / Framework: Theory explains causal relationships, Model provides a structured representation, and Framework organizes concepts within an analytical structure. Classification is based on semantic function rather than terminology.

4. Database / Information System: Databases are evaluated primarily in terms of content, coverage, and data quality, whereas Information Systems are evaluated in terms of functionality, interface design, retrieval performance, and usability. Classification depends on the evaluation perspective.

5. Policy / Policy Initiative: Policy refers to formal policy documents, plans, or regulations, whereas Policy Initiative refers to actions or measures undertaken during policy implementation. Policy actions should not be treated as policy documents.

### *Annotation standards and fallback strategy*

Both annotation and system outputs follow the principle of "most specific interpretable type first." The label "Other" is permitted only when a stable classification cannot be achieved after considering the context, ontology definitions, and boundary rules. Rather than serving as a substitute for proper classification, this label is intended to temporarily accommodate boundary cases. All instances assigned to "Other" must retain supporting evidence and documented sources of uncertainty, thereby facilitating subsequent ontology refinement, rule revision, and category expansion.

## Multi-Agent Collaborative Knowledge Mining System Architecture

Accordingly, this study proposes a multi-agent collaborative framework based on a candidate–validation–review paradigm for academic evaluation entity extraction. Given the stringent ontological constraints of the academic evaluation domain, allowing a single LLM to handle the entire workflow—including entity discovery, classification, and boundary identification—can easily amplify local errors throughout a single generation process (Han et al., 2024;  Du et al., 2024). Therefore, we decompose the task into three specialized Agents with clearly defined responsibilities, corresponding to high-recall candidate discovery, constraint-based validation, and domain-consistency review. This division is designed to improve the recall, accuracy, and overall consistency of extraction results. The output of each Agent is standardized as a structured intermediate representation that serves as the input for subsequent modules, rather than requiring the model to generate final extraction results from scratch.

### *Agent 1: Candidate Discovery*

Agent 1 is designed as a high-recall candidate discovery agent whose objective is to identify potential evaluation targets in academic evaluation texts as comprehensively as possible. Because the primary risk at this stage is candidate omission rather than misclassification, the system permits a moderate degree of over-generation and outputs identified candidates in a structured format for subsequent processing. To reduce invalid candidates, Agent 1 applies the principle of entity status to exclude expressions that lack independent academic semantic meaning or cannot be explicitly identified, thereby controlling noise while maintaining broad coverage. This stage does not perform final classification; instead, it aims to provide a comprehensive set of candidates for subsequent validation and ontology mapping.

### *Agent 2: Ontology-Constrained Validation and Classification*

Agent 2 is responsible for validating candidate entities and assigning hierarchical ontology-constrained categories. Unlike approaches that extract entities directly from raw text, Agent 2 operates on the candidate entities and supporting evidence generated by Agent 1, thereby restricting its role to validation and classification. First, the system determines whether each candidate represents a specific, identifiable academic evaluation object that can be mapped to an ontology category; candidates that fail to meet these criteria are discarded. Valid entities are then classified hierarchically according to ontology definitions, inclusion and exclusion rules, and the evaluative context in which they appear. This stage transforms high-recall candidate outputs into ontology-compliant structured annotations, reducing the risk that generic terms, abstract concepts, or evaluative expressions will be incorrectly recognized as entities.

### *Agent 3: Domain Review*

Agent 3 is designed as a domain review agent that independently re-examines extraction results after validation and classification have been completed. Drawing on entity categories, contextual evidence, and classification rationales, the agent checks for potential issues such as type drift, boundary inconsistencies, or insufficient supporting evidence and provides revision suggestions when necessary. In addition, Agent 3 generates an internal confidence assessment to assist the system in evaluating the reliability of extraction results. It should be noted that this confidence assessment serves solely as an internal reference signal; final performance evaluation is based on the silver-standard reference corpus described in the experimental design and validated through independent human review.

## EXPERIMENT DESIGN

### Research Questions

The experimental design of this study is structured around four research questions. RQ1 examines the overall effectiveness of the proposed framework relative to rule-based methods and direct-prompt single-model approaches. RQ2 investigates the contribution of the multi-agent workflow to candidate discovery, validity verification, and domain review. RQ3 evaluates whether ontology-based boundary constraints can improve fine-grained type classification and reduce confusion among neighboring entity types. RQ4 discusses the advantages and limitations of the complete framework under stricter typed entity evaluation criteria.

RQ1: Can the ontology-constrained multi-agent framework improve overall evaluation object extraction performance compared with rule-based baselines and direct-prompt zero-shot baselines?

RQ2: Can the multi-agent workflow improve extraction recall, output stability, and error traceability through task decomposition?

RQ3: Can ontology-based boundary constraints improve the accuracy of fine-grained type classification and reduce confusion among neighboring concepts?

RQ4: Under stricter typed entity evaluation criteria, what are the advantages and limitations of the complete framework?

To address these four research questions, this study proposes the Ontology-Guided Multi-Agent Entity Extraction System, as illustrated in Figure 1. The framework consists of four stages, which are described in detail in the following subsections.

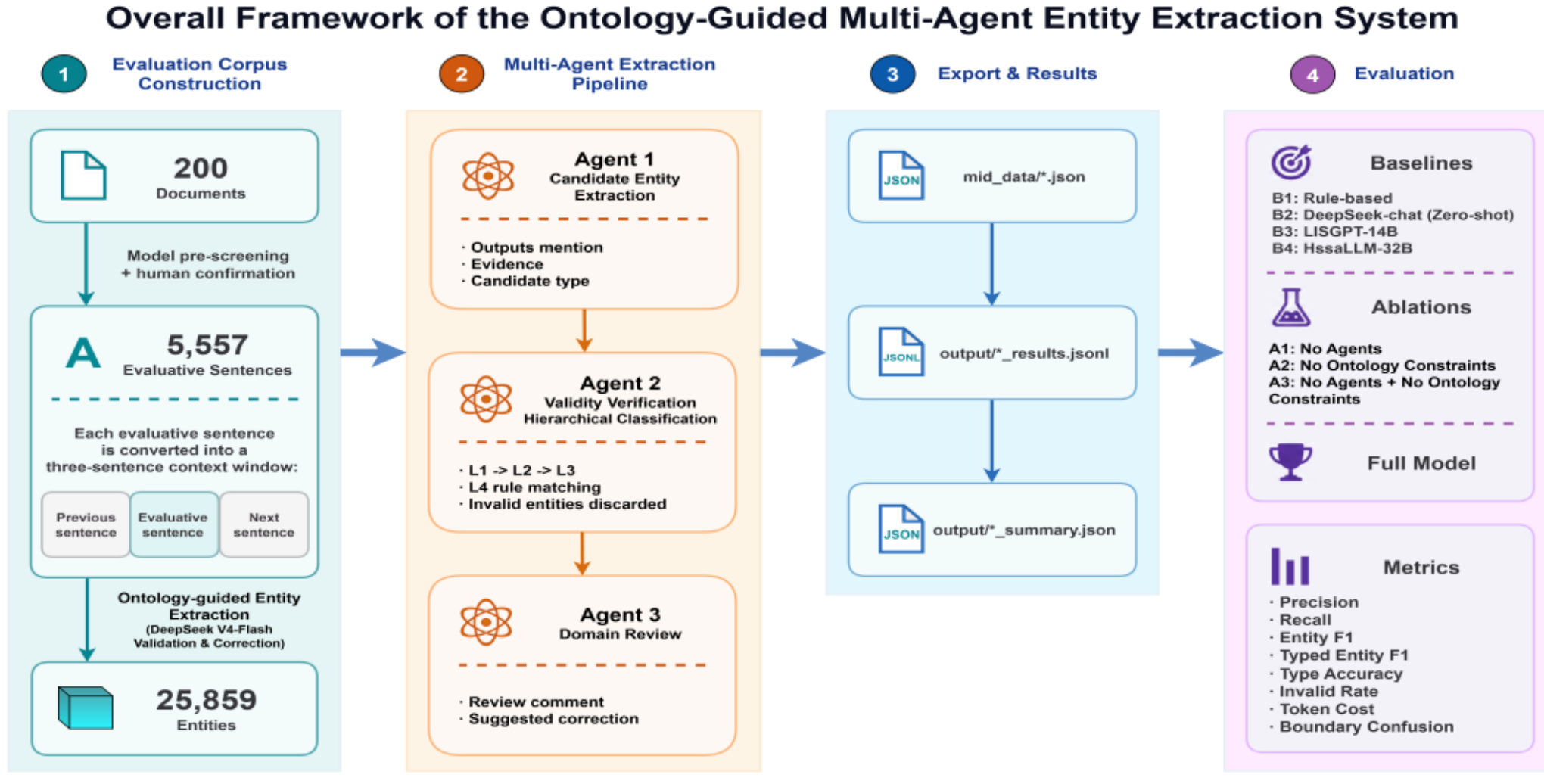

**Figure 1. Overall Framework**

### Corpus Construction and Validation

To construct the reference corpus, 200 Chinese Library and Information Science (LIS) review texts indexed in CSSCI were randomly selected. The DeepSeek-chat-based Full extraction framework was first applied to generate a high-recall candidate entity set. A separate model, DeepSeek V4-Flash, was then used only in the model-assisted corpus revision stage. Guided by ontology-constrained prompts, DeepSeek V4-Flash acted as an annotation reviewer rather than a primary annotator: it verified entity existence, boundary accuracy, and type assignments, removed false positives, supplemented missed entities, and generated the final revised entity set. Throughout the process, ontology definitions and boundary constraints were strictly enforced, requiring all entities to correspond to contiguous text spans and to be assigned the most specific contextually appropriate type whenever possible.

This procedure produced a silver-standard corpus containing 25,859 entities and 5,557 evaluative sentences, which was used as the reference corpus for subsequent experiments. To assess corpus quality, approximately 1.5% of all entities (388 entities) were randomly sampled and independently reviewed by two researchers. The inter-reviewer agreement was high (Cohen's $\kappa = 0.833$), while agreement between each reviewer and the silver standard exceeded 0.90 ($\kappa = 0.909$ and 0.910, respectively). Although a small number of boundary inconsistencies, redundant extractions, and type misclassifications were identified, the results indicate that the corpus is generally consistent with the ontology specifications and provides a reliable basis for subsequent evaluation. Because the reference corpus was constructed as a silver-standard corpus rather than a fully independent gold standard, the results should be interpreted as comparative evidence under ontology-constrained corpus construction. To reduce potential circularity, we conducted independent human validation on a random sample and reported agreement statistics as a quality check.

### Main Baselines and Ablation Settings

To ensure comparability, the main experiments evaluate only B1, B2, A1, A2, A3, and Full, all of which can be executed on the complete set of 200 test documents and share the same label space. All LLM-based settings (B2, A1, A2, A3, and Full) use the same backbone model (DeepSeek-chat) under identical decoding settings. B1 serves as the rule-based baseline, representing the lower performance bound of dictionary matching, regular expressions, and traditional named entity recognition (NER) approaches. B2 is a zero-shot baseline, which performs extraction and classification directly using a minimal prompt, a label list, and a predefined JSON output format. This baseline is designed to assess the direct adaptability of a general-purpose LLM in the absence of ontology-based boundary constraints, a multi-agent workflow, and internal validation mechanisms.

A1, A2, and A3 are internal ablation variants. A1 removes the multi-agent workflow while retaining type definitions, boundary rules, and structured output constraints. A2 preserves the multi-agent workflow but removes Ontology Constraints. A3 removes both the multi-agent workflow and Ontology Constraints, retaining only basic preprocessing, schema constraints, field validation, and minimal extraction instructions. B2 and A3 serve different purposes: the former is an external zero-shot prompting baseline, whereas the latter is a minimized ablation variant within the same engineering workflow. Therefore, the two settings are not intended for direct superiority comparisons.

Six experimental settings were designed to evaluate the contributions of different system modules:

· B1, a rule-based baseline using dictionary matching and regular expressions without LLM support, agent workflow, or ontology-based boundary constraints, representing the lower-bound performance of traditional methods.

· B2, a zero-shot baseline without agent workflow or ontology-based boundary constraints, representing direct prompting with a strong general-purpose LLM.

· A1, a few-shot setting with ontology-based boundary constraints but without agent workflow, used to evaluate the contribution of multi-agent task decomposition.

· A2, a few-shot setting with agent workflow but without ontology-based boundary constraints, used to evaluate the contribution of ontology-based boundary constraints.

· A3, a few-shot setting without agent workflow or ontology-based boundary constraints, used to assess performance degradation when both core mechanisms are removed.

· Full, the complete system combining few-shot prompting, multi-agent workflow, and ontology-based boundary constraints.

### Supplementary Domain Model Stability Analysis

In addition, B3 (LISGPT-14B) (Zhu et al., 2025) and B4 (HssaLLM-32B) (Zhao & Wang, 2026) were introduced as supplementary comparison models. The former one is a Chinese large language model developed for the Library and Information Science (LIS) domain and the latter one is for Humanities and Social Sciences. They are both included to examine the stability and performance differences between the proposed framework and representative domain-specific models. Because these models operate as independent systems and cannot be directly integrated into the proposed multi-agent framework, they are excluded from the main ablation analysis.

Two additional baselines were introduced to evaluate the stability of domain-specific LLMs under direct prompting:

· B3, a zero-shot setting based on LISGPT-14B with ontology-based boundary constraints but without agent workflow, used to evaluate the stability of direct prompting with a medium-scale domain-specific model.

· B4, a zero-shot setting based on HssaLLM-32B with ontology-based boundary constraints but without agent workflow, used to evaluate the stability of direct prompting with a larger domain-specific model.

### Evaluation Metrics

This study evaluates system performance from three perspectives: entity recognition, entity classification, and fine-grained distinction. Precision, Recall, and Entity-level F1 are used to assess entity extraction performance. Type Accuracy measures classification correctness among matched entities, while Strict Typed F1 serves as the primary strict metric requiring both entity boundaries and entity types to be correct. Invalid Rate and Intra-group Confusion are reported to reflect output validity and confusion among neighboring categories, respectively.

## RESULTS

### Overall Performance

RQ1 examines whether the ontology-guided multi-agent extraction framework improves evaluative target extraction compared with a shallow rule-based approach and a direct zero-shot LLM baseline. Table 1 compares B1, B2, and the full system on the same complete test set. B1 provides a lower-bound performance reference under limited semantic modeling capability, whereas B2 evaluates whether a strong general-purpose LLM can accomplish the task directly through a single-round zero-shot prompt.

| Setting | Precision | Recall | Entity-level F1 | Strict Typed F1 | Type Acc. | Invalid Rate |
|---|---|---|---|---|---|---|
| B1 | 40.93% | 20.67% | 27.47% | 13.10% | 47.68% | 0.00% |
| B2 | 73.07% | 65.07% | 68.84% | 46.99% | 68.26% | 9.97% |
| A1 | 82.89% | 64.81% | 72.74% | 57.41% | 78.92% | 7.94% |
| A2 | 80.08% | 74.03% | 76.94% | 56.60% | 73.57% | 11.19% |
| A3 | 75.63% | 74.07% | 74.84% | 53.51% | 71.51% | 6.63% |
| Full | 90.33% | 84.55% | 87.34% | 79.78% | 91.35% | 4.99% |

**Table 1. Results of Baseline and Ablation Experiments**

The full system substantially outperforms both baselines. Compared with B1, the full system achieves substantially higher precision and recall, while increasing Entity-level F1 from 27.47% to 87.34%, indicating that rule matching and conventional NER methods are insufficient for handling abstract evaluative targets. Compared with B2, the full system improves Entity-level F1 by 18.50 percentage points and Strict Typed F1 by 32.8 percentage points. The improvement in Strict Typed F1 is particularly important because it demonstrates that the performance gains arise not only from identifying more entities but also from assigning the correct L3 types to those entities.

### Role of the Multi-Agent Workflow and Ontology Constraints

RQ2 and RQ3 examine whether the two core mechanisms—the multi-agent workflow and ontology-based boundary constraints—make independent contributions to system performance. Table 1 also compares the full system with three ablation variants.

A1 removes the multi-agent workflow while retaining the ontology definitions and boundary rules. Its Entity-level F1 decreases to 72.74% and its Strict Typed F1 drops to 57.41%. This result indicates that compressing candidate discovery, classification validation, and domain review into a single prompting step substantially weakens both recall and strict type matching performance.

A2 retains the multi-agent workflow but removes the detailed ontology-based boundary constraints. Its Entity-level F1 reaches 76.94%, while Strict Typed F1 declines to 56.60% and Type Accuracy falls to 73.57%. These results suggest that agent specialization alone is insufficient to resolve fine-grained conceptual ambiguity. The system still requires explicit type definitions, inclusion and exclusion criteria, and guidance on commonly confused categories to constrain model judgments.

When both the multi-agent workflow and ontology-based boundary constraints are removed, A3 further declines to 74.84% Entity-level F1 and 53.51% Strict Typed F1. This finding indicates that the two mechanisms play complementary roles and jointly contribute to overall system effectiveness.

The difference between B2 and A3 should be interpreted as a diagnostic comparison. B2 reflects the direct prompting capability of a general-purpose LLM, whereas A3 represents the degraded performance of the proposed engineering workflow after its two core mechanisms have been removed.

### Error and Boundary Analysis

RQ4 investigates the remaining error patterns in the system and identifies where ontology-based boundary constraints contribute most substantially. Because the current experiments do not include Macro-F1 or L1-group F1, we conduct boundary error analysis using five predefined confusion groups defined in the Boundary Rules and Annotation Principles subsection of the Academic Evaluation Ontology. The analysis focuses on the most critical error categories in evaluative target extraction and those that most clearly demonstrate the effect of ontology-based boundary constraints.

As shown in Table 2, the full system maintains low confusion rates across all five predefined confusion groups, with particularly large reductions in the Methodology/Method/Technique group compared with A2 and A3. This finding suggests that the primary value of ontology-based boundary constraints lies in conceptually boundary-sensitive classification scenarios.

| Setting | Confusion Group 1 | Confusion Group 2 | Confusion Group 3 | Confusion Group 4 | Confusion Group 5 | Overall Confusion Rate |
|---|---|---|---|---|---|---|
| Full | 4.11% | 2.13% | 0.50% | 0.87% | 0.46% | 0.68% |
| A1 | 9.95% | 3.61% | 1.10% | 1.81% | 1.14% | 1.49% |
| A2 | 25.86% | 5.44% | 4.68% | 1.21% | 1.82% | 3.40% |
| A3 | 26.79% | 5.44% | 1.00% | 3.15% | 1.82% | 3.46% |

**Table 2. Results of Intra-group Confusion**

In many cases, the model correctly identifies entity boundaries but still misclassifies a methodology as a method, or vice versa. Consequently, Type Accuracy alone is insufficient for comprehensively evaluating system performance. Strict Typed F1 and Intra-group Confusion provide a more direct assessment of strict entity extraction quality by simultaneously reflecting boundary recognition and fine-grained type discrimination.

### Additional Analysis of Domain Model Performance

In the experimental design, B3 and B4 were introduced as supplementary references to contextualize the performance of the proposed framework against representative domain-specific models under direct prompting conditions. The results are presented in Table 3. To assess output stability, each setting was run three times, and the final results are reported as averages across runs.

| Model | Precision | Recall | Entity-level F1 | Type Acc. |
|---|---|---|---|---|
| Full | 90.33% | 84.55% | 87.34% | 91.35% |
| LISGPT-14B | 57.77% | 18.61% | 28.64% | 42.83% |
| HssaLLM-32B | 52.04% | 14.71% | 22.22% | 56.24% |

**Table 3. Results of Additional Analysis**

The supplementary results indicate that both domain-specific models exhibit relatively low recall under direct prompting conditions, suggesting that evaluative target extraction cannot be reduced to domain knowledge coverage alone. One possible explanation is that evaluative targets frequently span local contextual cues and ontology boundaries, requiring staged candidate generation and validation rather than a single-pass extraction process. In addition, entity recognition and extraction from academic review texts remains a relatively underexplored task, and the availability of high-quality task-specific training data is limited. Although domain-specific models demonstrate considerable future potential, the results suggest that they remain insufficiently adapted to this task under direct prompting settings.

## DISCUSSION

### Theoretical Contributions and Practical Implications

The theoretical contribution of this study lies in treating evaluation object extraction not merely as a named entity recognition task, but as a knowledge organization problem in scholarly evaluation. By defining evaluation objects as the semantic anchors to which evaluation dimensions and viewpoints can be attached, the proposed ontology provides a conceptual basis for transforming dispersed evaluative statements into structured evidence for research assessment and STI analysis.

The effectiveness of ontology-based boundary constraints lies in their ability to transform implicit disciplinary judgments into executable type boundaries. For frequently confused categories such as theory/model/framework, methodology/method/technique, and policy/policy initiative, the system no longer relies solely on the probabilistic generation capabilities of LLMs. Instead, it requires the model to make decisions based on functional attributes, epistemological status, and evaluative context. This explains the substantial advantages of Full over A2 in Type Accuracy, Strict Typed F1, and Intra-group Confusion.

The effectiveness of the multi-agent workflow does not stem from the use of multiple different models, but rather from organizing the reasoning process of the same backend model into an inspectable, staged workflow. Agent 1 prioritizes high recall and generates a candidate list; Agent 2 filters and classifies candidates under ontology-based boundary constraints; Agent 3 conducts domain review based on structured records. Compared with A1, which compresses all requirements into a single prompt, the advantages of Full arise from explicit intermediate states, error isolation, independent verification, and structured review. Candidate omissions, type confusions, and review disputes are no longer entangled within a single generation process but can instead be identified and corrected at different stages. This explains the superiority of Full over A1 in Recall, F1, and boundary control.

Practically, the framework offers a reproducible workflow for converting academic review texts, scholarly commentaries, and book reviews into structured evaluative evidence. Such evidence may support responsible research assessment, disciplinary mapping, and the development of STI platforms that move beyond citation-based indicators by incorporating expert judgment embedded in scholarly discourse.

### Limitations

Several limitations remain in this study. First, all three agents employ the same backend model in the main experiments. Therefore, this study does not claim to demonstrate the advantages of heterogeneous multi-model collaboration; rather, it validates the contributions of staged intermediate outputs, error isolation, independent verification, and structured review to evaluation object extraction. Second, the review scores generated by Agent 3 can only serve as internal diagnostic signals and cannot replace human annotation or external expert evaluation. In earlier results, weaker-performing models occasionally received higher Reviewer Confidence scores, indicating insufficient calibration in LLM self-assessment. Third, although the ontology was designed with the broader applicability of humanities and social sciences, the empirical validation in this study is limited to Chinese Library and Information Science review texts. The top-level categories and several boundary rules may be transferable to other HSS domains, but discipline-specific entity types, evaluative vocabularies, and boundary conditions still require further calibration and empirical testing in fields such as sociology, education, management, and history.

### Future Research

Future research may extend this work in two directions. First, the object layer can be transferred to additional disciplines to examine the generalizability and scalability of the ontology structure. Second, building upon the reliable identification of evaluation objects, future studies may further extract evaluation dimensions, evaluation viewpoints, and the semantic relationships among objects, dimensions, and viewpoints, thereby constructing a more comprehensive academic evaluation knowledge graph.

## CONCLUSION

Academic review texts, scholarly commentaries, and book reviews are not only important genres for knowledge dissemination and the organization of research trajectories, but also key vehicles through which academic communities conduct knowledge evaluation, quality assessment, and disciplinary positioning. As research evaluation increasingly emphasizes expert judgment, contextualized interpretation, and responsible assessment, the evaluative knowledge embedded in such texts has significant value for structured utilization. This study addresses the task of evaluation object extraction from Chinese academic review texts and proposes an ontology-constrained staged multi-agent extraction framework. The findings indicate that evaluation object extraction differs fundamentally from conventional named entity recognition. Its primary challenges lie in identifying abstract objects, determining context-dependent boundaries, and distinguishing between closely related entity types. To address these challenges, this study constructs an academic evaluation ontology centered on evaluation objects and decomposes the extraction process into three stages: candidate discovery, ontology-constrained classification, and domain review.

Experimental results show that, on the complete test set of 200 documents, the full system outperforms both the rule-based baseline and the single-model zero-shot baseline in Entity-level F1, Strict Typed F1, type accuracy, and boundary control. Supplementary experiments further indicate that direct prompting of domain-specific models still suffers from insufficient recall and output variability under small-sample, multi-seed settings, although these results are not intended as the primary basis for comparison with the full system. Ablation studies demonstrate that the multi-agent workflow primarily improves extraction stability and error traceability, whereas ontology-based boundary constraints mainly enhance fine-grained type classification and the discrimination of closely related concepts. Together, these two modules constitute the key mechanisms enabling the structured utilization of evaluative academic texts.

The contributions of this study are threefold. First, from a knowledge organization perspective, it conceptualizes evaluation object extraction from academic review texts as a knowledge structuring task governed by domain-specific semantic constraints. Second, it develops an object-level ontology tailored to evaluative contexts in Chinese Library and Information Science, with potential extension to broader humanities and social science domains. Third, it proposes a multi-agent extraction architecture centered on staged intermediate outputs, error isolation, independent validation, and structured review. Future research will focus on reducing inference costs, extending cross-disciplinary applicability, and further extracting evaluation dimensions, evaluation viewpoints, and their semantic relationships based on evaluation objects.

## GENERATIVE AI USE

We employed generative AI tools to assist with Python programming, language editing, text condensation, and formatting. All AI-generated outputs were reviewed, verified, and revised by the authors. The authors assume full responsibility for the research design, data analysis, interpretation, and content of this submission.

## ACKNOWLEDGMENTS

This article is an outcome of the major project "Research on the Construction of a Characteristic Chinese Evaluation System for Philosophy and Social Sciences from the Perspective of All-round Research Evaluation" (No. 24&ZD323) supported by the National Social Science Foundation of China, and the Postgraduate Research & Practice Innovation Program of Jiangsu Province titled "Optimization of the All-round Research Evaluation of Philosophy and Social Sciences Assisted by Generative AI".